\documentclass[11pt]{article}

\usepackage{acl}

\usepackage{times}
\usepackage{latexsym}
\usepackage[T1]{fontenc}
\usepackage[utf8]{inputenc}
\usepackage{microtype}
\usepackage{inconsolata}
\usepackage{graphicx}
\graphicspath{{figures/}}
\usepackage{booktabs}
\usepackage{tabularx}
\usepackage{array}
\usepackage{amsmath,amssymb,amsthm}
\usepackage{mathtools}
\usepackage{makecell}
\usepackage{tikz}
\usetikzlibrary{arrows.meta,positioning,calc,fit,shapes.multipart}
\usepackage{url}
\usepackage{xspace}
\usepackage{float}
\usepackage{stfloats}
\usepackage{algorithm}
\usepackage{algpseudocode}

\newcommand{\cedar}{\textsc{CEDAR}\xspace}
\newcommand{\voyager}{\textsc{Voyager}\xspace}
\newcommand{\codeact}{\textsc{CodeAct}\xspace}

\newcommand{\SigmaG}{\Sigma_{\mathrm{global}}}
\newcommand{\APact}{AP_{\mathrm{act}}}
\newcommand{\APobj}{AP_{\mathrm{obj}}}
\newcommand{\APevt}{AP_{\mathrm{evt}}}

\title{CEDAR: Automata as Verifiable Interfaces for Language-Guided Embodied Action}

\author{Lekai Chen, Alvaro Velasquez and Ashutosh Trivedi \\
  Department of Computer Science \\
  University of Colorado Boulder, USA \\
  \texttt{\{lekai.chen, alvaro.velasquez, ashutosh.trivedi\}@colorado.edu}}

\begin{document}
\maketitle
\begin{abstract}
Natural-language tasking of embodied agents is rarely just goal specification: users also impose constraints that must persist while the world changes. Code-generating LLM agents can produce plausible behaviors for such instructions, but their free-form programs provide no stable object to verify, compose with new constraints, or repair from a failing trace. We present \cedar, a counterexample-guided framework that grounds instructions as regular languages over environment event traces. \cedar uses a language model for semantic judgments and execution traces for correction, then represents both skills and specifications as deterministic finite automata. This turns constraints into executable finite-state objects: a learned skill can be intersected with a learned \textit{sleep at night} or \textit{stay in this biome} specification, yielding a controller that enforces the learned constraint by construction rather than by repeated prompting. In Minecraft, with the same simulator/API observations available to a program-generating baseline, \cedar maintains temporal and spatial constraints that the baseline fails to preserve and amortizes reuse of learned skills, reducing cumulative LLM queries. These results suggest that regular languages offer a practical verification layer between natural-language instructions and embodied-agent policies.
\end{abstract}

\section{Introduction}

Natural language is an attractive interface for tasking embodied agents, but instructions rarely describe only a terminal goal.  They also encode temporal, spatial, and safety constraints: \emph{collect diamonds but sleep at night}; \emph{explore the world but stay in the windswept forest}; \emph{mine minerals only at night}.  A useful agent must not merely execute a plausible program for the main goal; it must maintain a persistent interpretation of the user's constraints while acting in a changing environment.

Recent large-language-model (LLM) agents address open-world decision making by generating code, decomposing goals into subtasks, storing skills, or interleaving reasoning and action \citep{yao2023react,wang2023voyager,zhu2023ghost,liu2023agentbench,wang2024codeact}.  These systems are flexible, but their learned skills are usually free-form programs or text memories.  Such representations are hard to inspect formally.  A generated program may satisfy a constraint once and then forget it; two skills may be hard to compose safely; and a failure trajectory may not provide a precise object to repair.

\begin{figure*}[t!]
\centering
\includegraphics[width=\textwidth]{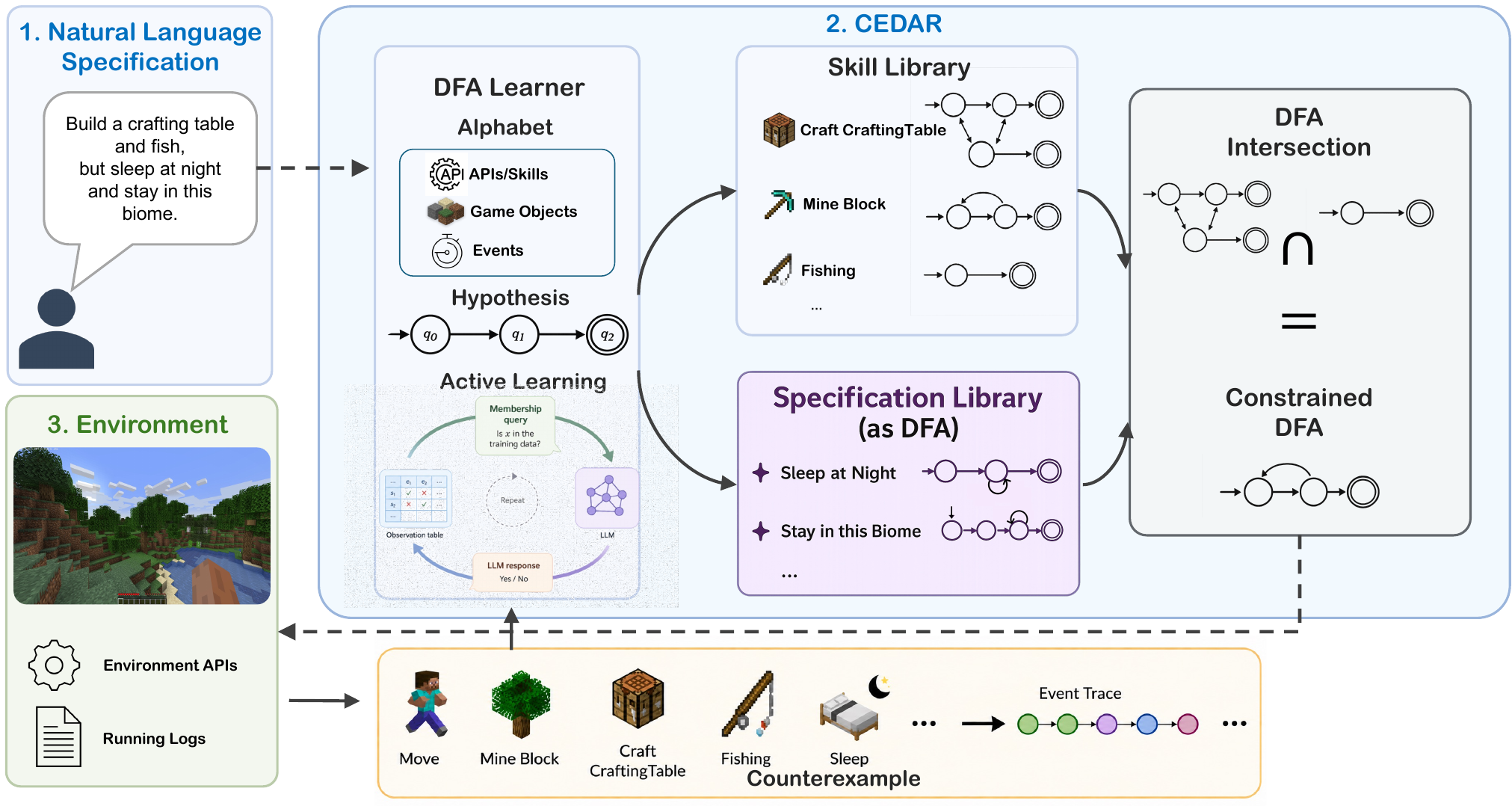}
\caption{\textbf{CEDAR workflow.} The system grounds a natural-language goal or constraint into a compact symbolic alphabet, learns skill and specification DFAs using LLM membership queries (MQs) and environment or human counterexamples (CEs), stores reusable skill automata, and verifies execution by DFA composition. EQ denotes equivalence-query feedback from execution or human inspection.}
\label{fig:pipeline}
\end{figure*}

This paper studies a different interface between language and action: grounding a natural-language instruction into a \emph{regular language over environment event traces}. Instead of treating the LLM as the final policy, we use it as a noisy semantic teacher. The LLM helps select a compact alphabet of action, object, and event-predicate symbols and answers membership queries about whether a trace is consistent with an instruction. A DFA learner builds an explicit automaton, while an environment-backed equivalence oracle tests the automaton and returns counterexamples. The resulting controller is less expressive than arbitrary code, but it is minimal up to isomorphism for the learned regular language, can be checked by replaying event traces, and is closed under operations such as intersection and concatenation.

We instantiate this idea in \cedar, a Counterexample-Driven Agent with Regular-language controllers. \cedar represents both \emph{skills} and \emph{specifications} as DFAs. Skill DFAs define executable progress toward goals, while specification DFAs encode constraints induced by natural language. A learned skill and specification can be combined through product intersection; prerequisite and follow-up skills can be combined through concatenation; and a failing execution trace can be returned to the learner as a counterexample. When a new task resembles a stored one, \cedar reuses the DFA through template adaptation over a substituted verb--object alphabet.

A common point of confusion is the word \emph{monitor}. In this paper, an event predicate or logger monitor is a read-only function over simulator/API state, such as time of day, biome, inventory, or whether an interaction changed the world. It is not a hidden perception module and not a hand-written safety shield. The code generated by \voyager{} has access to the same underlying state variables through the environment API. The difference is representational and architectural: \cedar converts natural-language goals and constraints into persistent finite-state objects that can be intersected, checked, reused, and repaired, whereas free-form code skills do not expose an formal language-level operation analogous to DFA intersection.

To our knowledge, \cedar is the first embodied-agent framework to use regular languages as a shared executable representation for both learned skills and learned natural-language constraints. This shared representation enables product intersection, concatenation, skill reuse under newly supplied constraints, and counterexample-guided relearning within one agent architecture. The main contribution is therefore not a new DFA-learning algorithm. CAPAL, active automata learning, and the use of an LLM as a semantic teacher build on prior work. The Minecraft experiments evaluate persistent constraint enforcement and amortized skill reuse. The iTHOR study evaluates only whether alphabet grounding and environment-backed counterexample repair can be instantiated in another simulator.

\section{Task Formulation}
\label{sec:formulation}

We formulate instruction following as language grounding over event traces.  Let $\mathcal{E}$ be an environment.  At each time step, the agent executes low-level actions and the environment exposes structured observations through its simulator/API: inventory, time, biome, and nearby objects.  \cedar uses deterministic logger predicates over these same observations to emit symbolic events.  A word extractor maps a trajectory of observations $\tau$ to a finite word $w = \phi(\tau) \in \Sigma^*$ over a task-specific alphabet $\Sigma$.

\paragraph{Alphabet.}
The global alphabet is a disjoint union
\begin{equation}
\SigmaG = \APact \uplus \APobj \uplus \APevt,
\end{equation}
where $\APact$ contains control primitives such as \texttt{mine}, \texttt{craft}, and \texttt{place}; $\APobj$ contains objects such as \texttt{diamond\_ore} or \texttt{bed}; and $\APevt$ contains event-predicate symbols such as \texttt{time=night}, \texttt{in\_biome(windswept\_forest)}, and \texttt{has\_$k$(item)}.  A symbol is therefore either an executable action-object event or an observed predicate emitted by the logger.  The logger predicate must be backed by an implemented environment query; the LLM selects among available predicates but does not create new sensors.

\paragraph{Instructions as languages.}
A user instruction $x$ induces two languages over traces: a goal language $L_g(x) \subseteq \Sigma^*$ and a constraint language $L_c(x) \subseteq \Sigma^*$.  For example, in \emph{collect diamonds but sleep at night}, the goal language accepts traces that reach diamond collection, while the constraint language rejects traces in which the agent continues working through night without sleeping.  \cedar learns DFA hypotheses $A_g$ and $A_c$ for these languages.  A DFA is $A=(Q,\Sigma,\delta,q_0,F)$, with states $Q$, transition function $\delta: Q \times \Sigma \to Q$, initial state $q_0 \in Q$, and accepting states $F \subseteq Q$.

\paragraph{Queries and counterexamples.}
We use active DFA learning algorithm CAPAL in a setting that is robust to persistent noise \citep{angluin1987learning,chen2026towards}.  A membership query (MQ) asks whether a word $w$ should be accepted.  In \cedar, an LLM answers MQs from the natural-language instruction and retrieved symbol descriptions.  An equivalence query (EQ) asks whether the current DFA matches the target.  For skill DFAs, the environment approximates the EQ oracle: the agent executes the DFA, and receives a CE if accepted behavior does not achieve the intended outcome or rejected behavior is feasible. For high-level specifications, a human may also provide CEs through annotation, demonstration, or direct trace judgment.

\begin{table}[t!]
\small
\centering
\begin{tabularx}{\linewidth}{lX}
\toprule
Term & Meaning in \cedar \\
\midrule
Alphabet $\Sigma$ & Task-specific set of actions, objects, and event-predicate symbols used to write traces. \\
Word $w$ & Sequence of symbols in $\Sigma$ extracted from environment logs. \\
Skill DFA & Automaton whose accepted words correspond to successful execution of a task skill. \\
Specification DFA & Automaton whose accepted words satisfy a persistent natural-language constraint. \\
MQ & Query asking whether a word is consistent with an instruction; answered by the LLM. \\
EQ/CE & Test of a hypothesis DFA; a mismatch yields a counterexample trace. \\
Regular restriction & The finite-state behavioral language imposed by a specification DFA; this is a learned formal constraint, not an extra sensor. \\
\bottomrule
\end{tabularx}
\caption{Language-grounding formulation.}
\label{tab:glossary}
\end{table}

\begin{figure*}[b!]
\centering
\includegraphics[width=\textwidth]{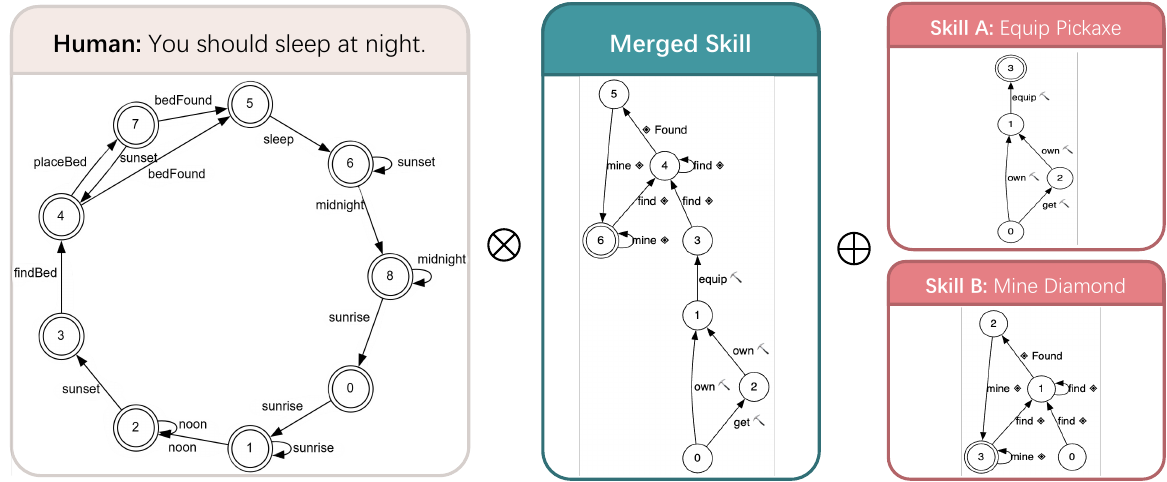}
\caption{\textbf{DFA composition for constraint enforcement.}
A learned specification DFA such as ``sleep at night'' is intersected
with a learned skill DFA such as ``mine diamond ore.'' The resulting
controller accepts exactly the traces accepted by both learned
automata. Rejecting sinks and nonessential transition detail are
omitted for readability.}
\label{fig:dfa-composition}
\end{figure*}

\section{CEDAR}
\label{sec:method}

The key design choice of \cedar is to separate semantic grounding from control structure.  The LLM supplies semantic judgments over language and symbols; the DFA supplies the persistent controller and formal operations.  This separation is what makes a natural-language constraint such as \emph{do not work at night} persist throughout execution rather than appearing only as a line of prompt context or a one-time generated subtask. Figure~\ref{fig:pipeline} summarizes \cedar.

\subsection{Alphabet Construction}
\label{sec:alphabet}

Learning over $\SigmaG$ is infeasible in an open-world environment with many APIs and objects, so \cedar constructs a task-specific sub-alphabet $\Sigma$ before DFA learning. The important assumption is not that \cedar receives extra predefined observations, but that the environment already exposes structured state through the same simulator/API used by the code-generation baselines. Each candidate symbol in $\SigmaG$ has two parts: a natural-language description used for retrieval and an executable binding to an environment query, action primitive, or logger predicate. The LLM may select and prune symbols, but it cannot create a sensor from raw pixels or use a predicate that lacks an implementation.

Our Minecraft global inventory contains 1,429 implemented symbols: 16 action primitives, 9 environment-event predicates, and 1,404 object symbols. The alphabet constructor embeds the instruction, retrieves candidate verbs, objects, and event predicates from this inventory, and asks the LLM to prune them into the finite alphabet used by the active learner. Evaluation against target alphabets from 44 validated skill DFAs yields absolute accuracy $0.9372\pm0.10$ and overlap coefficient $0.9208\pm0.10$. The prompt and complete retrieval analysis are reported in Appendix~\ref{app:alphabet-details}.

If a required predicate is not implemented in $\SigmaG$, \cedar cannot express the intended regular language, just as a code-generating agent cannot reliably condition on an unavailable simulator variable. If the predicate exists globally but is omitted from the retrieved sub-alphabet, execution may expose the omission when the resulting behavior conflicts with an outcome recognized by the approximate equivalence oracle. In that case, \cedar expands the alphabet and restarts learning. This mechanism is not complete: an omitted predicate may prevent the learned DFA from representing part of the instruction while the available oracle nevertheless accepts the resulting behavior. The DFA can then reach an accepting state despite failing to capture the full intended semantics.

\subsection{Active DFA Learning under Noise}
\label{sec:active-learning}

For a fixed instruction and alphabet, the learner repeatedly asks MQs, updates a DFA hypothesis, and tests it through execution.  The LLM is not assumed to be a perfect oracle.  We follow the probabilistic MAT view and use the CAPAL persistent-noise-aware active DFA-learning procedure \citep{chen2026towards}: repeated and inconsistent MQ answers are reconciled in the query cache, while environment CEs correct systematic semantic errors. The system-level instantiation, including cache resolution, hypothesis construction, equivalence testing, is given in Appendix~\ref{app:capal-pseudocode}.

For skill learning, the EQ oracle is implemented by execution. The current DFA is converted into a policy by choosing a shortest path from the current automaton state to an accepting state. If an invoked transition is absent from logs or fails to produce the expected event, the edge is temporarily disabled and the path is recomputed. If no accepting path remains, the trace is stored as a negative CE. If the DFA reaches an accepting state but the environment does not satisfy the intended goal, the accepted trace is likewise returned as a CE. The CE is passed to CAPAL's refinement routine, which updates the learner's observation storage and reconstructs the next hypothesis. Learning terminates when execution succeeds without producing a CE within the interaction budget.

\subsection{Skill Storage, Retrieval, and Adaptation}

A learned skill is stored as
\begin{equation}
  s = \langle A, v, n, E, D \rangle,
\end{equation}
where $A$ is the DFA, $v \in \APact$ is a verb, $n \in \APobj$ is an object, $E \subseteq \APevt$ is the set of success events, and $D \subseteq \Sigma^* \times \{0,1\}$ is the labeled evidence used to learn the DFA.  The evidence set is the union of MQ labels and EQ counterexamples.

For a new query $(v',n')$, the Skill Manager first attempts exact symbolic lookup.  If $v'=v$ and $n'=n$, the stored DFA is reused directly.  If the verb matches but the object differs, the manager treats the existing DFA as a template: it substitutes noun-specific symbols, updates examples accordingly, and refines the adapted DFA through the active learner.  If the verb differs, the system learns a new skill.  This symbolic reuse is the main source of amortized efficiency.

\subsection{Verifying Natural-Language Constraints}
\label{sec:constraint-composition}

Specifications are learned as DFAs with the same procedure, except that human counterexamples may be used when the intended constraint is ambiguous.  The specification DFA is the persistent constraint object: it is learned from natural language and counterexamples, not hand-coded as an external guard.  Once a specification DFA $A_c$ and a skill DFA $A_g$ are available, \cedar enforces their conjunction by product intersection:
\begin{equation}
  L(A_g \cap A_c) = L(A_g) \cap L(A_c).
\end{equation}
The composed automaton accepts exactly the traces accepted by both learned automata. This is an exact guarantee with respect to $A_g$ and $A_c$; it is not independently a certificate that either learned automaton faithfully captures every clause of the original natural-language instruction. Appendix~\ref{app:state-counts} reports the component-state counts, the Cartesian-product upper bounds, etc. A code-generating baseline can be prompted with the same natural-language constraint, and in our experiments it is; however, its stored skills are free-form programs rather than formal languages, so there is no exact operation analogous to $L(A_g)\cap L(A_c)$ for reusing a previously learned skill under a newly supplied constraint. Section~\ref{sec:experiments} separately tests whether adding the same runtime assertion-and-repair condition to \voyager{} and \codeact{} closes this gap.

\cedar also chains skills through concatenation.  If $A_1$ accepts traces for a prerequisite skill and $A_2$ accepts traces for a follow-up skill, accepting states of $A_1$ are connected to the initial state of $A_2$.  Unlike unconstrained sequential calls to code skills, the DFA composition exposes whether the first skill actually reached an accepting condition before the second is invoked.

\begin{figure*}[t!]
\centering
\includegraphics[width=\textwidth]{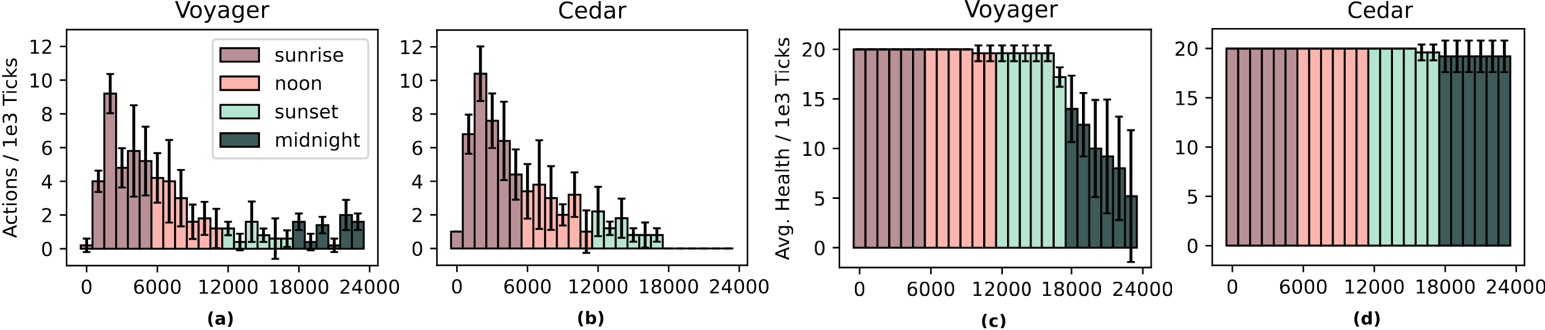}
\caption{\textbf{Sleep-at-night constraint.} Action counts and average health for \voyager{} and \cedar{} over repeated trials lasting three Minecraft days (aggregated over 24,000 ticks). Health denotes the averaged exact health value in the gameplay. Curves and error regions denote means and standard deviations. \cedar{} suppresses night-time work because the constraint remains active throughout execution.}
\label{fig:sleep-at-night}
\end{figure*}

\begin{figure*}[t!]
\centering
\includegraphics[width=\textwidth]{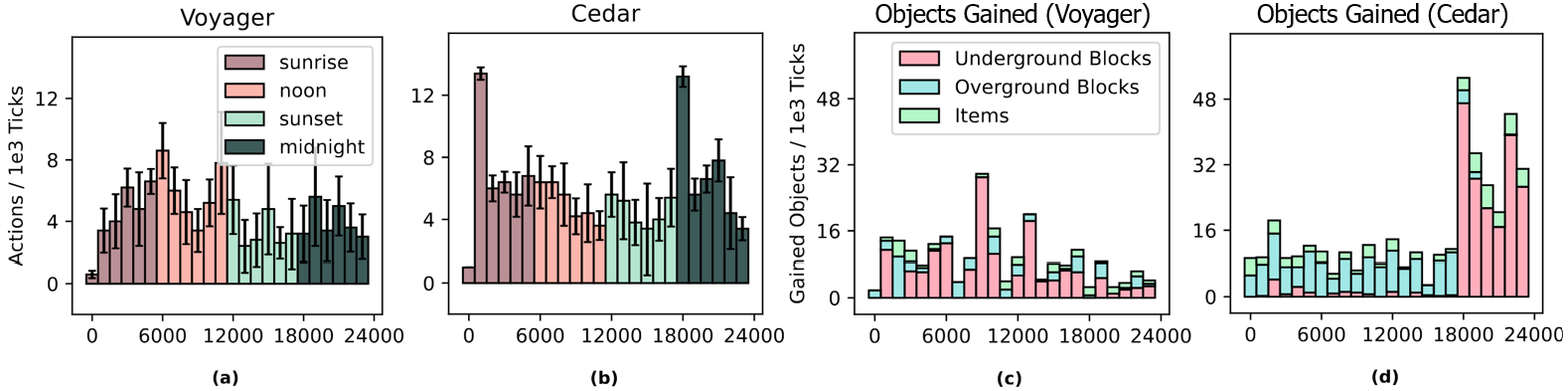}
\caption{\textbf{Mine-minerals-only-at-night constraint.}
Temporal distribution of actions and collected objects over five
trials per method, with a fixed budget of 24,000 simulator ticks per
trial. The direct compliance metric---the fraction of mineral-mining
actions executed outside the night interval---is reported in
Table~\ref{tab:mine-night}.}
\label{fig:mine-at-night}
\end{figure*}

\section{Experiments}
\label{sec:experiments}

We evaluate four questions.  \textbf{Q1:} Under matched simulator/API observations, can regular-language controllers maintain natural-language constraints during execution?  \textbf{Q2:} Does DFA skill reuse improve goal completion and LLM query cost?  \textbf{Q3:} Are the LLM-grounded alphabet and counterexample loop reliable enough for practical learning?  \textbf{Q4:} Can the same grounding pipeline be instantiated outside Minecraft?

The primary environment is Minecraft. We compare against \voyager{}~\citep{wang2023voyager} and \codeact{}~\citep{wang2024codeact}, including versions of each method with and without an assertion-and-repair mechanism. All methods receive the same task instruction and access to the same underlying simulator/API state variables. For the \voyager{} comparisons, we use the same map, spawn location, simulator APIs, and task instruction. \cedar's event predicates are deterministic recodings of state variables available through those APIs and do not provide additional perceptual information.

Human CEs are not used for task-completion skill learning, the code-baseline comparisons, or the iTHOR experiments. They are used only in the high-level specification demonstrations, for which Appendix~\ref{app:human-ce-accounting} reports the feedback modality, number of interventions, annotation source, and cost. We evaluate five iTHOR tasks as a feasibility study of alphabet grounding and environment-backed CE repair, not as a comparative benchmark or as evidence of cross-simulator skill reuse or specification intersection. The LLM components use GPT-5.5 for task decomposition and MQs, GPT-5.4-mini for JSON translation, and \texttt{text-embedding-3-large} for embeddings.

\subsection{Constraint Following in Minecraft}

\paragraph{Sleep at night.} The instruction is: \emph{craft a diamond pickaxe and keep collecting diamonds; sleep at night; you are given a bed}. We evaluate \voyager{} and \codeact{} both with and without the same assertion \texttt{night -> sleep before controller yield}, and compare them with \cedar's learned specification-DFA composition. Consequently, a reused or long-running code skill may continue executing after night begins until control returns to the assertion boundary.Table~\ref{tab:assertion-baselines} reports task completion, constraint compliance, joint success, and health over five trials per condition. A run is counted as successful if it finds a diamond, as constraint-compliant if it doesn't violate any assertions programmed by experts, and as jointly successful only if both events occur in the same trajectory. Health is computed as terminal health at corresponding trajectory ticks.

\begin{table*}[t!]
\small
\centering
\begin{tabular}{lcccc}
\toprule
Method
& Found diamond
& Constraint compliant
& Joint success
& Mean health $\uparrow$ \\
\midrule
\voyager{} without assertions
& 4/5 & 0/5 & 0/5 & $4.0\pm3.16$ \\
\voyager{} with assertions
& 4/5 & 2/5 & 0/5 & $8.4\pm4.56$ \\
\codeact{} without assertions
& 3/5 & 1/5 & 0/5 & $4.8\pm3.03$ \\
\codeact{} with assertions
& 0/5 & 3/5 & 0/5 & $12.0\pm5.83$ \\
\textbf{\cedar{}}
& \textbf{5/5}
& \textbf{5/5}
& \textbf{5/5}
& $\mathbf{18.4\pm2.61}$ \\
\bottomrule
\end{tabular}
\caption{\textbf{Sleep-at-night assertion comparison.}
Exact 95\% binomial confidence intervals are reported in Appendix~\ref{app:assertion-statistics}. Because $n=5$, the comparison is interpreted descriptively rather than as a high-powered estimate of population-level success probabilities.}
\label{tab:assertion-baselines}
\end{table*}

Runtime assertions improve local compliance, but none of the four code
conditions achieves joint task-and-constraint success in the five
evaluated trials. Trace inspection indicates that the assertions can repair a violation only when control returns to the assertion boundary: plans may treat sleep as a one-time subgoal, and reused or long-running skills may continue after night begins. \cedar instead includes the learned specification state in the composed controller throughout execution.

\paragraph{Mine minerals only at night.}
The instruction is: \emph{explore the world and collect as many different items as possible, but dig for minerals like iron and diamond only at night}. Both methods receive a fixed budget of 24,000 simulator ticks per trial. We measure constraint compliance through the daytime mineral-mining action counts.  Table~\ref{tab:mine-night} summarizes object collection and Figure~\ref{fig:mine-at-night} shows the temporal distribution of actions and collected objects.  \cedar{} performs more total actions, collects more underground and overground objects under the reported protocol, and obtains more total items.  The result reflects the difference between a one-time subtask and a persistent temporal restriction: \cedar's higher action count reflects its use of daytime for exploration and nighttime for mineral extraction under the fixed horizon.

\begin{table}[t!]
\small
\centering
\resizebox{\linewidth}{!}{%
\begin{tabular}{lccccc}
\toprule
Method & Actions & Underground & Overground & Items & Objects \\
\midrule
\voyager{} & $106\pm4.6$ & $157\pm15$ & $48.6\pm8.7$ & $24.8\pm6$ & $229\pm14$ \\
\cedar{} & $137\pm12$ & $189\pm30$ & $141\pm17$ & $57.4\pm4.7$ & $387\pm13$ \\
\bottomrule
\end{tabular}}
\caption{Minecraft instruction following for ``mine minerals only at night.'' Values are mean $\pm$ standard deviation for the corresponding recorded events.}
\label{tab:mine-night}
\end{table}

\begin{figure*}[t!]
\centering
\includegraphics[width=0.82\textwidth]{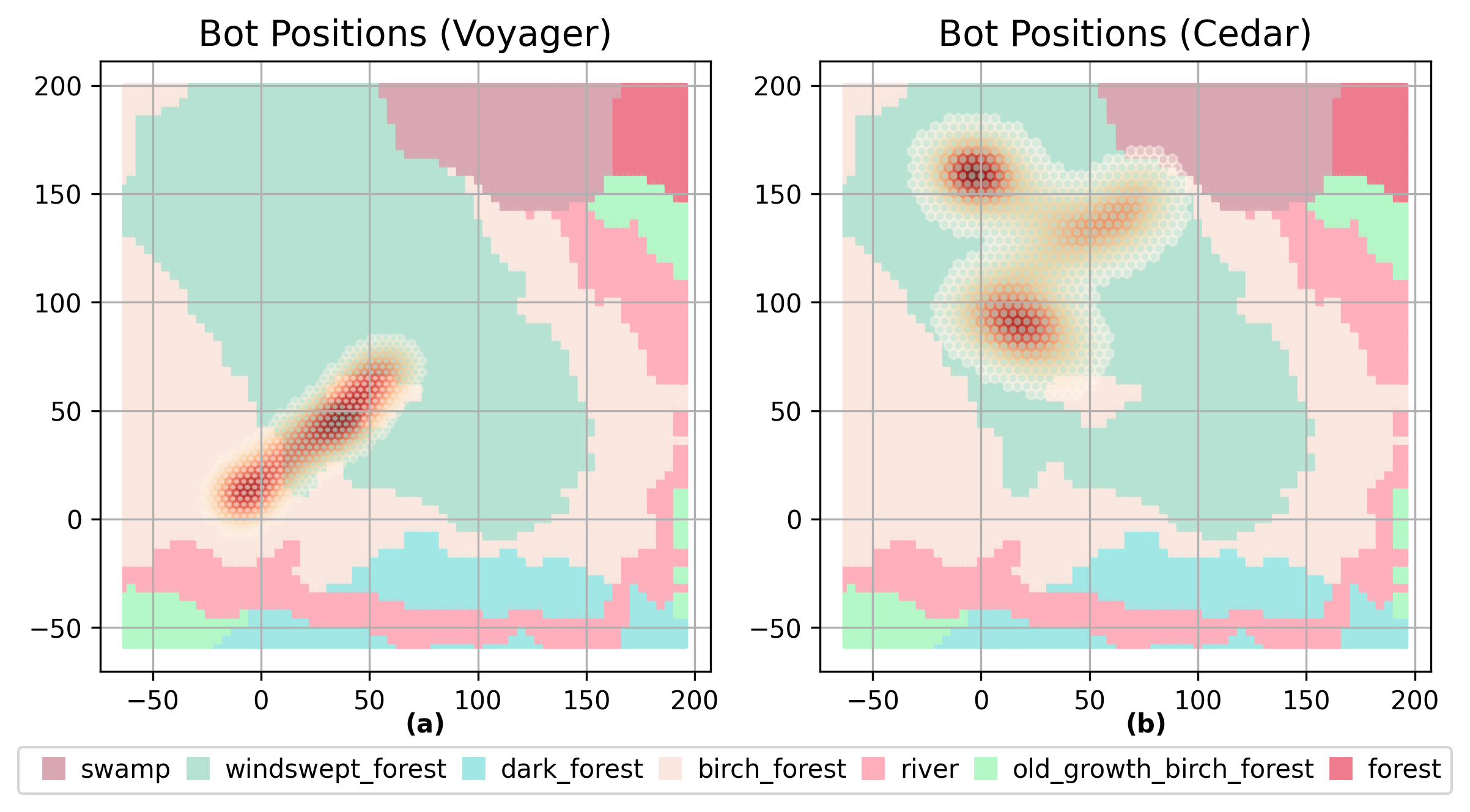}
\caption{\textbf{Biome constraint.} Background colors denote Minecraft biomes and heatmap intensity denotes bot activity. \cedar{} remains in the requested windswept forest region, while \voyager{} moves across biome boundaries. The heatmaps show an aggregation of location visit frequencies over 5 trials.}
\label{fig:biome}
\end{figure*}

\paragraph{Stay in a requested biome.}
For the instruction \emph{explore the world but stay in the windswept forest}, \cedar{} includes biome-predicate symbols in the alphabet and learns a spatial constraint DFA.  The biome predicate is obtained from the same simulator state available to both agents; \cedar{} differs by turning it into a persistent formal constraint.  The baseline traverses multiple biomes, while \cedar{} restricts activity to the target biome.  Figure~\ref{fig:biome} shows the resulting activity heatmaps.  This experiment shows that the same language-grounding mechanism handles non-temporal event predicates.

\subsection{Goal Completion and Skill Reuse}

Table~\ref{tab:goal-completion} reports prompting iterations and success rates for Minecraft crafting tasks. Without a skill library, \cedar{} incurs additional interactions because it must learn and test a DFA. With a learned library, \cedar{} adapts stored DFAs instead of repeatedly prompting for new code. Among successful trajectories, \cedar{} with a skill library uses fewer mean prompting iterations than \voyager{} with a skill library on four of the five tasks. These means are conditioned on success, however, and therefore summarize different trajectory subsets when success rates differ. They are not unconditional efficiency estimates. Appendix~\ref{app:success-stats} reports exact confidence intervals for the success rates, while Appendix~\ref{app:query-runtime} provides the cleaner cumulative-query accounting for the amortization claim.

\begin{table*}[t!]
\small
\centering
\resizebox{\textwidth}{!}{
\begin{tabular}{lccccc}
\toprule
Method & Wooden Pickaxe & Iron Pickaxe & Diamond Pickaxe & Lava Bucket & Compass \\
\midrule
\voyager{} w/o S.L. & $6.9\pm2.1$ (25/25) & $30.4\pm4.5$ (25/25) & $33.4\pm10.4$ (11/25) & $26.4\pm7.2$ (19/25) & $25.9\pm2.9$ (17/25) \\
\voyager{} & $4.2\pm3.5$ (25/25) & $16.2\pm3.8$ (25/25) & $28.9\pm10.7$ (17/25) & $22.3\pm5$ (25/25) & $17.7\pm2.8$ (25/25) \\
\cedar{} w/o S.L. & $6.6\pm2.8$ (25/25) & $30.4\pm3.3$ (25/25) & $45.9\pm11.1$ (18/25) & $27.2\pm5.6$ (25/25) & $29.8\pm2.1$ (9/25) \\
\cedar{} & $5.5\pm4$ (25/25) & $9.6\pm6$ (25/25) & $19.8\pm6.5$ (23/25) & $10.5\pm8.5$ (25/25) & $10\pm2.7$ (25/25) \\
\bottomrule
\end{tabular}
}
\caption{\textbf{Minecraft goal completion.} Each method--task condition contains 25 trials. Entries report prompting iterations as mean $\pm$ standard deviation over successful trajectories only. Exact 95\% binomial confidence intervals for success rates are reported in Appendix~\ref{app:success-stats}. S.L. denotes a skill library.}
\label{tab:goal-completion}
\end{table*}

Appendix~\ref{app:query-runtime} reports cumulative LLM-query and wall-clock accounting for the amortization claim; these resource-budget measurements are kept separate from statistical task-performance estimates.

\subsection{Grounding Reliability}  The alphabet and CCE loop are explicit dependencies of the method, so we analyze them separately from end-to-end task success. Against their validated target alphabets, retrieval achieves absolute accuracy $0.9372\pm0.10$ and overlap coefficient $0.9208\pm0.10$. Appendix~\ref{app:alphabet-details} reports the complete retrieval analysis. Appendix~\ref{sec:ce-discover} separately evaluates environment-backed CE discovery by perturbing correct skill DFAs. The average reported discovery accuracy is 0.9358. Together, these analyses support the practical use of an LLM semantic teacher corrected through environment feedback while leaving absent, rare, or noisy predicates as central failure modes. In particular, neither analysis establishes that every instruction-relevant predicate will be retrieved or that a silent alphabet omission will always be detected.

\subsection{iTHOR Feasibility Study}
\label{sec:ithor-feasibility}

\begin{figure*}[t!]
\centering
\includegraphics[width=\textwidth]{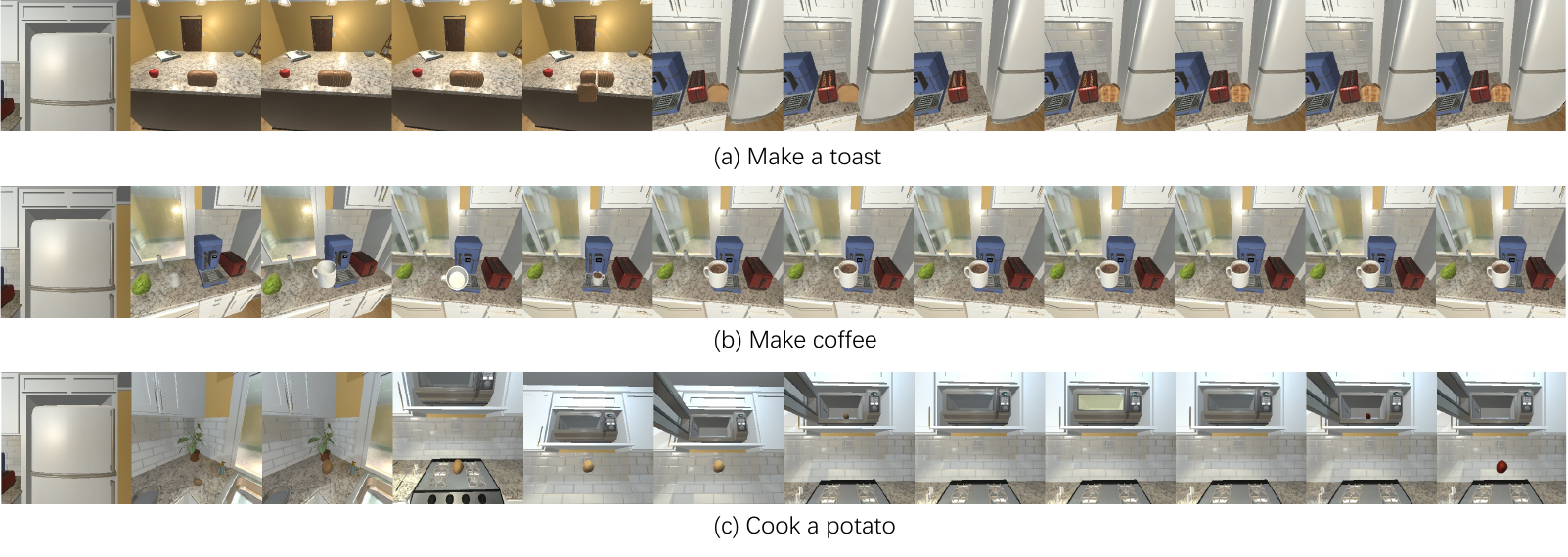}
\caption{\textbf{iTHOR grounding-and-repair feasibility rollouts.}Time-lapse frames for making toast, making coffee, and cooking a potato. The potato task initially produced a CE due to camera perspective; adding a \texttt{LOOKUP} action repaired the DFA without extra LLM queries. These rollouts show that the pipeline can run outside Minecraft, but they are not used as a comparative benchmark. Results are based on 3 runs.}
\label{fig:ithor}
\end{figure*}

To test whether task-specific alphabet grounding and
environment-backed counterexample repair can be instantiated outside
Minecraft, we run five iTHOR tasks: \textsc{Make Toast},
\textsc{Make Coffee}, \textsc{Cook Potato}, \textsc{Make Salad}, and
\textsc{Store Plate in Fridge}. They do not evaluate persistent
constraint enforcement, skill reuse, or amortized reuse outside
Minecraft. Each task uses task-specific alphabet construction, executable
predicate bindings, and an execution-backed counterexample loop.
Each skill requires two LLM calls: one for sub-alphabet construction
and one to synthesize the code-based oracle program that binds logger
predicates and action stubs. Four tasks succeed in a single pass.
\textsc{Cook Potato} initially fails because of camera perspective;
the approximate equivalence oracle returns a counterexample, and
inserting a \texttt{LOOKUP} action repairs the DFA without additional
LLM calls. Representative rollouts are shown in
Figure~\ref{fig:ithor}. We do not present \voyager{} results in iTHOR
because its released implementation is coupled to Minecraft and
Mineflayer APIs. Accordingly, all comparative claims, persistent
constraint-enforcement claims, and skill-reuse claims in this paper
are supported by the Minecraft evaluation.

\section{When Do Regular Languages Suffice?}
\label{sec:regularity}

A central limitation of \cedar{} is also its source of interpretability: DFAs recognize regular languages.  They cannot count indefinitely, store unbounded memory, or implement arbitrary recursion as JavaScript programs can.  We therefore characterize the intended use case.

Let $\Sigma$ be a task alphabet extracted from implemented logger predicates, including thresholded predicates such as \texttt{has\_$3$\_cobblestone}, time predicates such as \texttt{time=night}, and spatial predicates such as \texttt{in\_biome(x)}. If all numerical conditions needed for a skill are exposed as finite milestone predicates and each subtask completes or fails within a bounded event horizon $H$, then the successful traces of that subtask are regular. Intuitively, these predicates convert environment state into finite symbols, and the bounded horizon yields a finite control graph. Under these assumptions, intersection for constraint enforcement and concatenation for skill chaining preserve regularity.

This analysis clarifies both the power and the boundary of the approach.  \cedar{} is appropriate when instructions can be expressed through finite milestones and persistent temporal or spatial restrictions backed by available environment predicates.  It is less appropriate when the task requires unbounded counting, arbitrary data structures, or long-range memory not captured by the logger.  In such settings, DFAs may still be useful as learned constraints, but the skill representation should be extended to richer formal models such as counter automata, visibly pushdown automata, or recursive state machines.

\section{Related Work}
A related line translates natural-language instructions into temporal logic~\citep{fuggitti2023nl2ltl,pan2023data,liu2024lang2ltl}. These methods typically assume a proposition vocabulary and study formula translation using paired natural-language and formal specifications. \cedar{} instead retrieves an API-backed task alphabet and learns an executable DFA through active judgments and execution CE. The supervision assumptions are therefore different, and we do not empirically establish superiority over NL-to-LTL translation systems. An NL-to-LTL front end could initialize a specification DFA when its output is regular or compilable to a finite monitor; subsequent CE-guided repair would remain complementary.

Automata have previously been used as a verification layer for LLM-generated robot programs. In particular, \citet{yang2024joint} convert generated programs into automaton-based representations and verify them against externally supplied safety specifications. This establishes the broader use of automata for verifying generated programs, which we do not claim as novel. \cedar{} addresses a different interface: it learns both the skill automaton and the natural-language constraint automaton from semantic judgments and execution traces, stores both as executable agent representations, and then composes the learned objects.

Classical active DFA learning identifies regular languages through membership and equivalence queries~\citep{angluin1987learning}. Recent work studies LLMs as probabilistic semantic teachers and persistent-noise-tolerant active DFA learning \citep{chen2026towards}; related work also learns automata from natural-language oracles~\citep{vazquez2024lstar}. \cedar{} inherits these learning ideas rather than introducing a new active automata-learning algorithm. Its contribution is their use within an embodied-agent architecture in which both reusable skills and learned language constraints share the same executable regular-language representation.

\citet{xu2020joint} jointly infer reward machines and reinforcement- learning policies by collecting episodes whose observed rewards are inconsistent with a hypothesis reward machine and then relearning the hypothesis. \cedar{} likewise uses execution inconsistencies as counterexamples, but addresses a different setting and interface: it constructs task-specific alphabets from simulator APIs, uses an LLM to judge natural-language trace membership, executes learned DFAs as controllers, and composes learned skill and specification automata. Counterexample-guided automaton refinement itself is therefore inherited rather than claimed as novel. Safe reinforcement learning and shielding constrain policies during learning or execution~\citep{moos2022robust,gu2024safe}. These methods generally assume that the relevant constraint or shield is already available. \cedar{} addresses the preceding grounding problem by learning a finite-state specification from natural language and counterexamples. A learned \cedar{} specification DFA could in principle be used as a shield around another controller, but this paper does not compare against systems supplied with independently validated formal specifications and does not claim that its learned DFAs are certified safety specifications. More broadly, recent work on efficient LLM reasoning reduces the effective reasoning space through relevance-based pruning, structured search, model routing, and cache reuse \citep{ye-etal-2026-tableqa,guo-etal-2026-rethinking-table,ye2026rethinkingstepwisemodelrouting,wen-etal-2026-speccache}; CEDAR is complementary in reducing the symbolic decision space through task-specific alphabet retrieval.

\section{Conclusion}  We presented \cedar, the first framework that uses regular languages as a verifiable representation for learned embodied skills and natural-language constraints. Within the evaluated symbolic, event-driven setting, this shared representation makes learned instruction semantics explicit: skill and specification automata can be combined by product intersection, skills can be chained and reused, and failing execution traces can be returned as counterexamples for relearning. The Minecraft experiments provide evidence for persistent temporal and spatial constraint following, including relative to \voyager{} and \codeact{} variants augmented with the same runtime assertion-and-repair condition. They also provide the evidence for amortized skill reuse. The iTHOR study demonstrates that the same pipeline can be instantiated in another simulator. Overall, the results support regular languages as a practical common substrate for learned skills and constraints when instructions can be represented through finite predicates.

\section*{Limitations}
Evidence for persistent constraint enforcement, skill composition, and amortized reuse comes from Minecraft. The iTHOR study evaluates only alphabet grounding, executable predicate binding, and environment-backed counterexample repair. CAPAL assumes structured simulator/API observations and does not reliably extract predicates from raw visual input. The evaluated environments are discrete and symbol-rich; we claim no advantage over code or continuous-control policies requiring unbounded memory, arbitrary data structures, or fine-grained continuous state.

DFA intersection exactly enforces the languages represented by the learned automata, but fidelity to the user's instruction depends on the selected alphabet and the correctness of membership- and equivalence-query feedback. Missing predicates can silently omit intended clauses while still permitting acceptance. Execution-backed alphabet expansion may reveal some omissions, but it is incomplete, and we do not systematically measure their rate.

Incorrect human counterexamples can define the wrong learning target and steer CAPAL toward an incorrect automaton. Appendix~\ref{app:human-ce-accounting} documents the human-feedback demonstrations, but this study does not establish robustness to incorrect or conflicting feedback. Where only one provider was used, inter-annotator sensitivity cannot be estimated.

Product automata have worst-case multiplicative state growth. Although the tested compositions remain small (Appendix~\ref{app:state-counts}), this does not guarantee general scalability. Retrieval is evaluated over the current 1,429-symbol inventory without controlled variation of inventory size or target-automaton complexity.

Finally, learned specification DFAs are not certified safety specifications. They are insufficient for physical robots or other safety-critical systems without independently validated sensing, specifications, runtime enforcement, and fail-safe control. Exact composition cannot correct incorrectly grounded semantics.

\section*{Acknowledgments}
This work was supported by DARPA award number HR0011261E001 and the NSF under CAREER Award CCF-2146563. Ashutosh Trivedi is a Royal Society Wolfson Visiting Fellow and gratefully acknowledges support from the Wolfson Foundation and the Royal Society.

\bibliography{references}

\clearpage
\appendix

\section{LLM Query and Runtime Cost}
\label{app:query-runtime}

\begin{figure}[htbp]
\centering
\includegraphics[width=\linewidth]{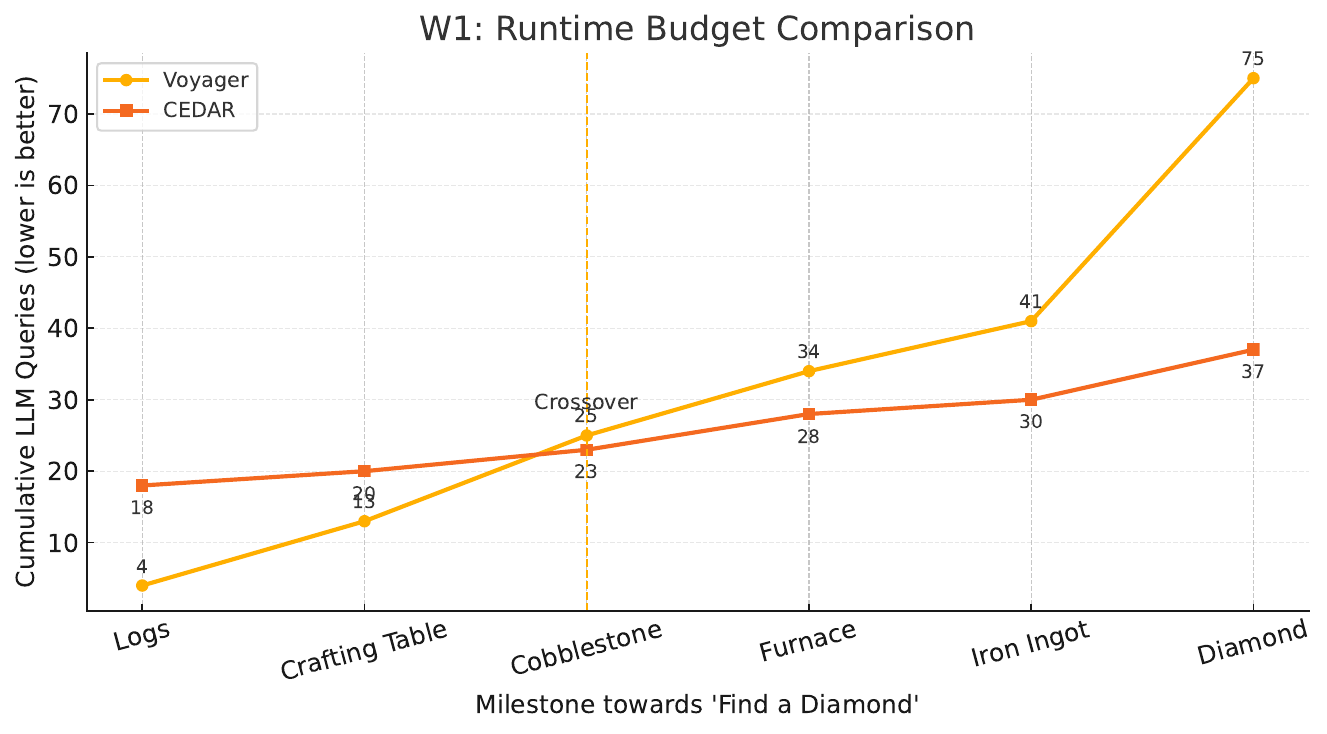}
\caption{\textbf{Cumulative LLM-query budget} for the \emph{Find a Diamond} milestone sequence. \cedar{} pays an early alphabet-construction cost but amortizes it through DFA reuse and reaches the final milestone with fewer total queries.}
\label{fig:query-budget}
\end{figure}

The cost of DFA learning is front-loaded.  Once a DFA is learned, retrieval, intersection, and concatenation are symbolic operations.  On the \emph{Find a Diamond} milestone sequence, \cedar{} uses more queries at the first logging milestone but fewer cumulative queries from the Cobblestone milestone onward, reaching the final diamond milestone with 37 LLM queries compared with 75 for \voyager{} (Table~\ref{tab:queries} and Figure~\ref{fig:query-budget}).  Wall-clock measurements show the same amortization effect: \cedar{} incurs DFA construction time but has much lower skill-addition and total execution time under the reported protocol (Table~\ref{tab:runtime}).  These resource-budget results are separate from statistical task-performance estimates: success or object-count claims should be supported by confidence intervals, while amortization claims are based on cumulative LLM-query and runtime accounting.

\begin{table}[htbp]
\small
\centering
\resizebox{\linewidth}{!}{%
\begin{tabular}{lrrrrrr}
\toprule
Method & Logs & Table & Cobble & Furnace & Iron & Diamond \\
\midrule
\voyager{} & 4 & 13 & 25 & 34 & 41 & 75 \\
\cedar{} & 18 & 20 & 23 & 28 & 30 & 37 \\
\bottomrule
\end{tabular}}
\caption{Cumulative LLM queries on the \emph{Find a Diamond} milestone sequence.  Lower is better.}
\label{tab:queries}
\end{table}

\begin{table}[H]
\small
\centering
\resizebox{\linewidth}{!}{%
\begin{tabular}{lcc}
\toprule
Stage & \voyager{} & \cedar{} \\
\midrule
Task decomposition & $2.749\pm1.238$ & $3.943\pm2.134$ \\
Code/sample generation & $6.381\pm1.990$ & $5.548\pm1.876$ \\
Program description & $2.384\pm1.208$ & N/A \\
Skill addition & $2.653\pm1.228$ & $0.021\pm0.008$ \\
DFA construction & N/A & $18.548\pm14.289$ \\
Skill retrieval & $0.323\pm0.235$ & $0.089\pm0.586$ \\
Total execution & $62.427\pm55.834$ & $33.101\pm24.391$ \\
\bottomrule
\end{tabular}}
\caption{Stage-wise wall-clock time in seconds, reported as mean $\pm$ standard deviation over 5 independent measurements per stage.}
\label{tab:runtime}
\end{table}

\section{Alphabet Retrieval Details}
\label{app:alphabet-details}

\begin{figure}[H]
\centering
\includegraphics[width=\linewidth]{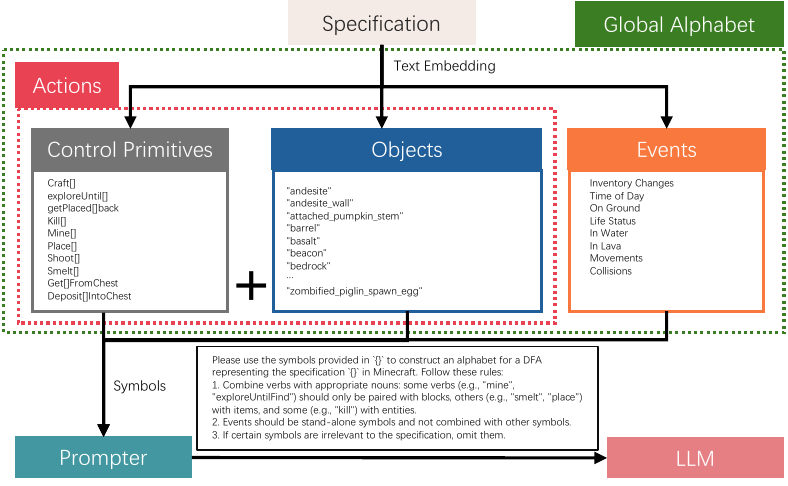}
\caption{\textbf{RAG-based alphabet construction.} From the global action, object, and event-predicate alphabet, retrieval selects task-relevant symbols that the LLM prunes into the finite alphabet used for DFA learning.  The selected predicates are symbolic recodings of existing simulator/API observations, not additional perceptual inputs.}
\label{fig:rag-alphabet}
\end{figure}

The alphabet constructor retrieves candidate symbols with implemented environment bindings and asks the LLM to build a compact alphabet.  We evaluate the retriever against target alphabets derived from 44 validated skill DFAs.  Table~\ref{tab:rag} reports high absolute accuracy and overlap, indicating that the LLM usually reasons over a relevant subalphabet rather than the full global space.  This is an analysis of symbol retrieval, not a claim that \cedar invents new sensors.

\begin{table}[H]
\small
\centering
\resizebox{\linewidth}{!}{%
\begin{tabular}{lccc}
\toprule
Metric & Absolute Acc. & Overlap Coef. & Cosine Sim. \\
\midrule
RAG & $0.9372\pm0.10$ & $0.9208\pm0.10$ & $0.4500\pm0.14$ \\
\bottomrule
\end{tabular}}
\caption{RAG alphabet-construction performance over 44 validated skill DFAs.}
\label{tab:rag}
\end{table}

\section{Additional Results}

\subsection{iTHOR Query Budget}

\begin{table}[t]
\small
\centering
\resizebox{\linewidth}{!}{%
\begin{tabular}{lcc}
\toprule
Task & LLM calls & Outcome \\
\midrule
\textsc{Make Toast} & 2 & success \\
\textsc{Make Coffee} & 2 & success \\
\textsc{Cook Potato} & 2 & success after CE fix \\
\textsc{Make Salad} & 2 & success \\
\textsc{Store Plate in Fridge} & 2 & success \\
\bottomrule
\end{tabular}}
\caption{iTHOR query budget for the feasibility study.  The counterexample fix for \textsc{Cook Potato} did not require additional LLM calls. Each task is evaluated over 3 executions.}
\end{table}

\subsection{Counterexample Discovery by Item}
\label{sec:ce-discover}

\begin{table}[H]
\small
\centering
\begin{tabular}{lcc}
\toprule
Item & Accuracy & Standard deviation \\
\midrule
Dirt & 0.9727 & 0.1629 \\
Birch Log & 0.8636 & 0.3432 \\
Grass Block & 1.0000 & 0.0000 \\
Birch Leaves & 0.9909 & 0.0949 \\
Stone & 0.9727 & 0.1629 \\
Coal Ore & 1.0000 & 0.0000 \\
Iron Ore & 1.0000 & 0.0000 \\
Copper Ore & 0.9909 & 0.0949 \\
Gold Ore & 0.9636 & 0.1872 \\
Redstone Ore & 0.9636 & 0.1872 \\
Emerald Ore & 0.4909 & 0.4999 \\
Diamond Ore & 0.9818 & 0.1336 \\
Lapis Ore & 0.9636 & 0.1872 \\
Andesite & 0.9818 & 0.1336 \\
Granite & 0.9636 & 0.1872 \\
Sand & 0.8727 & 0.3333 \\
Average & 0.9358 & 0.1692 \\
\bottomrule
\end{tabular}
\caption{Counterexample discovery in Minecraft simulations after perturbing correct skill DFAs.}
\label{tab:ce-discover}
\end{table}

\begin{figure}[H]
\centering
\includegraphics[width=\linewidth]{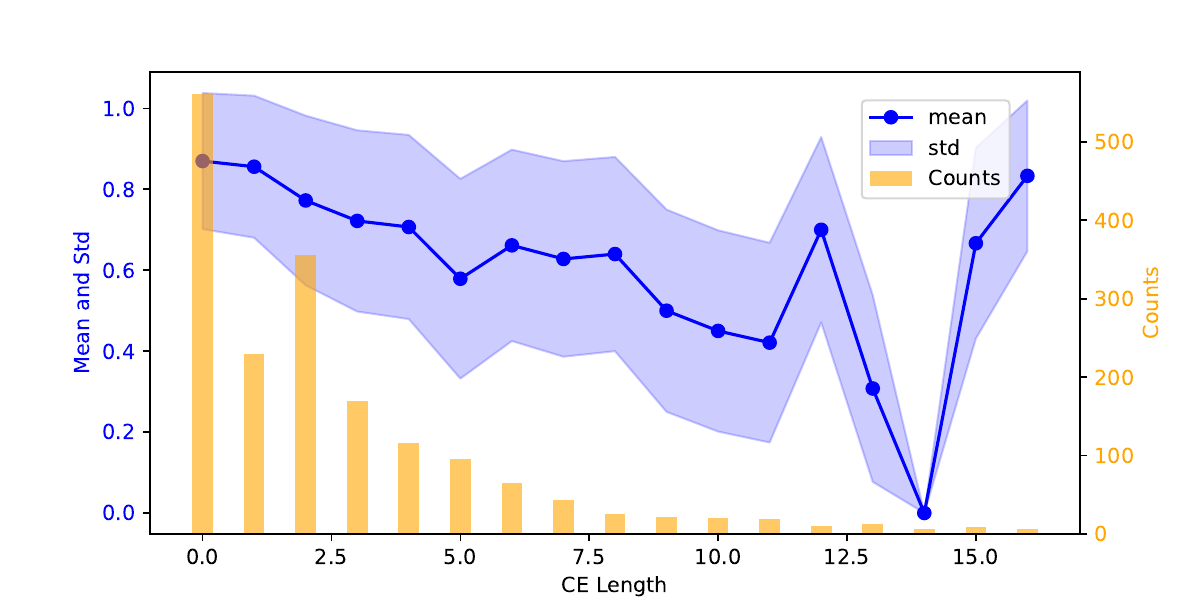}
\caption{\textbf{Counterexample collection by CE length.} Shorter counterexamples are easier to surface during simulation; most observed CEs have length below seven, where collection probability remains high.}
\label{fig:env-ce}
\end{figure}

\subsection{Expert DFA Reference}

As an upper-bound diagnostic for the representation, manually designed DFA skills were tested on the same Minecraft tasks.  This is not a learning baseline: it assumes expert construction and bypasses the language-grounding problem.  The expert DFAs achieved 10/10 success for Wooden Pickaxe, Iron Pickaxe, Lava Bucket, and Compass, and 9/10 for Diamond Pickaxe, with the failure caused by map initialization.

\section{Human Counterexample Accounting}
\label{app:human-ce-accounting}

Human feedback is used only for the high-level specification
demonstrations. It is not used for Minecraft task-completion skill
learning, the code-baseline comparisons, or the iTHOR experiments.
Table~\ref{tab:human-ce-accounting} reports the interventions actually
used in the reported specification-learning runs.

\begin{table}[t]
\small
\centering
\resizebox{\linewidth}{!}{%
\begin{tabular}{lcc}
\toprule
Specification & Human CEs & Final DFA states \\
\midrule
Sleep at night
  & 8 & 9 \\
Mining only at night
  & 5 & 5 \\
Stay in windswept forest
  & 4 & 2 \\
\bottomrule
\end{tabular}}
\caption{Human counterexamples used for specification-DFA learning.
The modality identifies trajectory annotation, demonstration logs, or
direct formal trace judgment. Time reports the total annotation time
used in the reported run rather than a theoretical intervention bound.}
\label{tab:human-ce-accounting}
\end{table}

The reported specification feedback was supplied by
on human expert (author). Consequently, the experiments do not measure inter-annotator disagreement or sensitivity to alternative interpretations. A semantically incorrect counterexample can define an incorrect target language and cause CAPAL to learn the wrong specification DFA.

\begin{figure}[H]
\centering
\includegraphics[width=\linewidth]{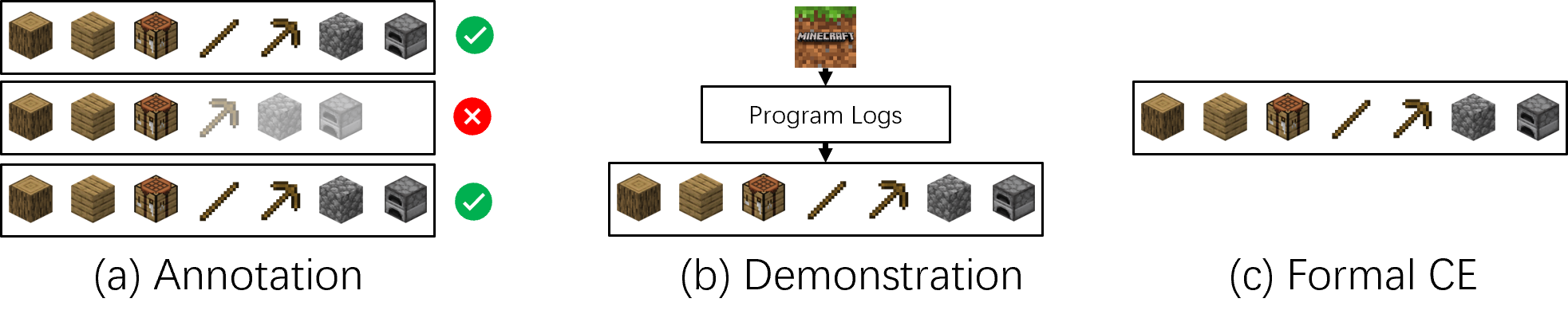}
\caption{\textbf{Human counterexample interfaces.}
Counterexamples can be supplied through trajectory annotation,
demonstration logs, or direct formal words over the DFA alphabet.}
\label{fig:human-ce}
\end{figure}

\section{Prompt Sketch for Alphabet Construction}

The RAG-style alphabet constructor retrieves candidate symbols with implemented environment bindings and asks the LLM to build a compact alphabet.  A simplified prompt is shown below:

\begin{quote}\small
Given the specification and the candidate symbols in braces, construct an alphabet for a DFA in Minecraft.  Combine verbs with appropriate nouns: \texttt{mine} should be paired with blocks, \texttt{smelt} and \texttt{place} with items, and \texttt{kill} with entities.  Event predicates should remain standalone symbols.  Omit irrelevant symbols.  Return only the selected symbols.
\end{quote}

\section{Failure Modes}

\paragraph{Missing predicates.}
If no implemented symbol represents a required milestone, the DFA
cannot encode it. For example, a task requiring at least three
cobblestones needs a finite event predicate such as
\texttt{has\_3\_cobblestone}. Without that predicate, the DFA may
over-accept or fail to distinguish relevant progress states. If the
predicate exists in the global inventory but is omitted from the
retrieved task alphabet, an execution counterexample may expose the
omission and trigger alphabet expansion. This recovery is not
guaranteed: if the available oracle does not recognize the semantic
mismatch, the incomplete DFA may still reach an accepting state. This
is a logger/alphabet limitation rather than an observation-advantage
assumption, since code-generating agents also require access to the
relevant simulator variable.

\paragraph{Rare long-horizon counterexamples.}
Counterexamples become harder to discover when an incorrect transition appears only after a long execution prefix.  Most simulated counterexamples are short, but rare long-horizon failures remain possible.

\paragraph{Shared semantic misunderstanding.}
The LLM and human feedback provider may share an incorrect
interpretation of an ambiguous instruction. A semantically incorrect
counterexample can then define the wrong target language and direct
the active learner toward an incorrect automaton. Counterexamples help
only when at least one oracle can recognize the mismatch. \cedar{}
does not claim robustness to systematically incorrect equivalence
feedback.

\section{CEDAR Instantiation of CAPAL}
\label{app:capal-pseudocode}

Algorithm~\ref{alg:cedar-capal} gives the system-level interaction between CEDAR and CAPAL. CAPAL maintains its active-learning state, including its prefix set, suffix sets, state partition, discrimination tree, and hypothesis DFA. CEDAR supplies semantic membership judgments, persistent query-cache evidence, and execution-backed counterexamples.

\begin{algorithm}[H]
\caption{CEDAR instantiation of CAPAL}
\label{alg:cedar-capal}
\begin{algorithmic}[1]
\Require Instruction $x$; initial alphabet $\Sigma$; CAPAL learner
         $\mathcal{L}$; membership-query budget $B_{\mathrm{MQ}}$;
         execution/EQ budget $B_{\mathrm{EQ}}$
\Ensure DFA hypothesis $A$ or budget-exhaustion report
\State Initialize persistent MQ cache $C \gets \varnothing$
\State Initialize gold counterexample cache $G \gets \varnothing$
\State Initialize $\mathcal{L}$ on $\Sigma$
\State $b_{\mathrm{MQ}} \gets 0$; $b_{\mathrm{EQ}} \gets 0$
\While{\textbf{true}}
    \While{\Call{HasPendingMQ}{$\mathcal{L}$}}
        \State $w \gets \Call{NextMQ}{\mathcal{L}}$
        \If{$w \in G$}
            \State $\hat{y} \gets G[w]$
        \ElsIf{$w \in C$}
            \State $\hat{y} \gets C[w]$
        \Else
            \If{$b_{\mathrm{MQ}} = B_{\mathrm{MQ}}$}
                \State \Return budget-exhaustion report and latest hypothesis,
                       if available
            \EndIf
            \State $\hat{y} \gets \Call{LLMJudge}{x,\Sigma,w}$
            \State $C[w] \gets \hat{y}$
            \State $b_{\mathrm{MQ}} \gets b_{\mathrm{MQ}}+1$
        \EndIf
        \State \Call{AnswerMQ}{$\mathcal{L},w,\hat{y}$}
    \EndWhile

    \State $A \gets \Call{BuildHypothesis}{\mathcal{L}}$
    \If{$b_{\mathrm{EQ}} = B_{\mathrm{EQ}}$}
        \State \Return budget-exhaustion report and $A$
    \EndIf
    \State $(\tau,o) \gets \Call{Execute}{A}$
    \State $b_{\mathrm{EQ}} \gets b_{\mathrm{EQ}}+1$
    \State $w_{\tau} \gets \phi(\tau)$

    \If{\Call{OutcomeMismatch}{$x,A,w_{\tau},o$}}
        \State $y_{\tau} \gets
               \Call{EnvironmentLabel}{x,w_{\tau},o}$
        \State $G[w_{\tau}] \gets y_{\tau}$
        \State $\mathcal{L} \gets
               \Call{RefineWithCounterexample}
               {\mathcal{L},w_{\tau},y_{\tau}}$
    \ElsIf{\Call{DetectedMissingPredicate}{$x,\Sigma,\tau,o$}}
        \State $\Sigma \gets
               \Call{ExpandAlphabet}{x,\Sigma,\tau,o}$
        \State Reinitialize $\mathcal{L}$ on the enlarged $\Sigma$
        \State Replay compatible cached labels from $C$ and $G$
    \Else
        \State \Return $A$
    \EndIf
\EndWhile
\end{algorithmic}
\end{algorithm}

\paragraph{Persistent-noise reconciliation.}
For each queried word $w$, the cache stores the first Boolean membership judgment $\widetilde{y}(w)$ and returns that judgment on subsequent queries. CAPAL does not repeatedly sample the same word or aggregate multiple labels for that word. A counterexample label supplied through execution is instead stored as a gold override and takes precedence over the cached membership judgment.

CAPAL obtains robustness through a statistical class query over multiple suffixes. For prefixes $u$ and $v$, let
\[
\mathcal{E}
=
E_{\mathrm{core}}\cup E_{\mathrm{pool}}',
\qquad
m=|\mathcal{E}|,
\]
where $E_{\mathrm{pool}}'$ is a capped sample from the suffix pool. CAPAL computes the empirical disagreement rate
\[
D(u,v)
=
\frac{1}{m}
\left|
\left\{
e\in\mathcal{E}:
\widetilde{y}(ue)\neq\widetilde{y}(ve)
\right\}
\right|
\]
and declares $u$ and $v$ to represent the same Myhill--Nerode class iff
\[
D(u,v)\leq p_0+\tau,
\qquad
p_0=2\bar{\eta}(1-\bar{\eta}),
\]

\[
\tau=
\min\left\{
\tau_{\max},
\sqrt{\frac{\ln(2/\alpha)}{2m}}
\right\},
\]
where $\bar{\eta}$ is a conservative upper bound on the membership-noise rate. Only \textsc{Different} outcomes are cached. The cache key includes the prefix pair, suffix-set version, and decision threshold, so a comparison is recomputed when its evidence or threshold changes. During a comparison with $m$ planned suffixes, CAPAL stops early and caches \textsc{Different} when the accumulated disagreement count $d_t$ satisfies
\[
\frac{d_t}{m}>p_0+\tau.
\]

\paragraph{Counterexample update.}
CAPAL maintains a prefix-closed access set $S$, a core discriminator set $E_{\mathrm{core}}$, a capped suffix pool $E_{\mathrm{pool}}$, an observation table of cached membership judgments, and a discrimination tree over the current state classes. Given a labeled counterexample $(w_{\tau},y_{\tau})$, CAPAL stores $y_{\tau}$ as a gold override and inserts the prefixes of $w_{\tau}$ into $S$.

CAPAL then applies Rivest--Schapire analysis, searching for the earliest factorization
\[
w_{\tau}=uae
\]
such that
\[
\delta^{A}(A[u],a)\neq A[ua].
\]
If such a factorization exists, the suffix $e$ is added to $E_{\mathrm{core}}$ as a structural discriminator. If no split exists, the counterexample is treated as label-only: its gold label corrects the acceptance decision for the class it reaches, and the counterexample is recorded without immediately expanding the observation table. If the same label-only counterexample recurs, its suffixes are added to $E_{\mathrm{core}}$ to separate the incorrectly merged class.

When enforcing consistency, CAPAL uses the discrimination tree to select a discriminator from the lowest common ancestor of two distinct successor classes and adds the corresponding derived column to $E_{\mathrm{core}}$. Any membership queries required by the resulting closedness or consistency repairs are issued lazily. CAPAL then rebuilds its state partition and hypothesis DFA; CEDAR does not directly patch an individual transition.

\paragraph{Budgets and termination.}
The membership and execution budgets in Algorithm~\ref{alg:cedar-capal} are experiment-level CEDAR limits rather than parameters required by CAPAL. Cache hits and gold-label lookups do not consume the membership-query budget; only a new call to \textsc{LLMJudge} does. Each execution of a hypothesis consumes one execution/EQ query.

Learning terminates when execution produces no recognized counterexample, when the experiment's success predicate is satisfied, or when either budget is exhausted. CAPAL ordinarily assumes a fixed alphabet. Alphabet expansion is therefore implemented as a CEDAR-level restart: the alphabet-dependent learner state is reinitialized, while cached membership judgments and gold counterexample labels for words that remain valid over the enlarged alphabet are replayed. This preserves previously acquired semantic evidence without reusing an invalid state partition or discrimination tree.

\section{Assertion-Baseline Statistics}
\label{app:assertion-statistics}

Table~\ref{tab:assertion-ci} reports exact two-sided 95\% binomial
confidence intervals for the three binary outcomes in
Table~\ref{tab:assertion-baselines}. The intervals are necessarily
wide because each condition contains only five trials.

\begin{table*}[t]
\small
\centering
\begin{tabular}{lccc}
\toprule
Method
& Found diamond
& Constraint compliant
& Joint success \\
\midrule
\voyager{} w/o A.
& \makecell{4/5\\$[0.284,0.995]$}
& \makecell{0/5\\$[0.000,0.522]$}
& \makecell{0/5\\$[0.000,0.522]$} \\

\voyager{} w/ A.
& \makecell{4/5\\$[0.284,0.995]$}
& \makecell{2/5\\$[0.053,0.853]$}
& \makecell{0/5\\$[0.000,0.522]$} \\

\codeact{} w/o A.
& \makecell{3/5\\$[0.147,0.947]$}
& \makecell{1/5\\$[0.005,0.716]$}
& \makecell{0/5\\$[0.000,0.522]$} \\

\codeact{} w A.
& \makecell{0/5\\$[0.000,0.522]$}
& \makecell{3/5\\$[0.147,0.947]$}
& \makecell{0/5\\$[0.000,0.522]$} \\

\cedar{}
& \makecell{5/5\\$[0.478,1.000]$}
& \makecell{5/5\\$[0.478,1.000]$}
& \makecell{5/5\\$[0.478,1.000]$} \\
\bottomrule
\end{tabular}
\caption{Observed success counts and exact two-sided 95\%
Clopper--Pearson confidence intervals for the five-trial
sleep-at-night comparison. The intervals describe marginal binary
outcomes and do not remove the uncertainty caused by the small sample. A.: \texttt{aseertions}.}
\label{tab:assertion-ci}
\end{table*}

\section{Goal-Completion Success Intervals}
\label{app:success-stats}

Table~\ref{tab:goal-success-ci} reports exact two-sided 95\% binomial
confidence intervals for the success counts in
Table~\ref{tab:goal-completion}. These intervals concern task success;
they do not address the conditional prompting-iteration means.

\begin{table}[!ht]
\scriptsize
\centering
\setlength{\tabcolsep}{2.5pt}
\renewcommand{\arraystretch}{1.08}

\begin{tabular}{lcccc}
\toprule
Task
& \makecell{\voyager{}\\w/o S.L.}
& \voyager{}
& \makecell{\cedar{}\\w/o S.L.}
& \cedar{} \\
\midrule

Wooden
& \makecell{25/25\\$[0.863,1.000]$}
& \makecell{25/25\\$[0.863,1.000]$}
& \makecell{25/25\\$[0.863,1.000]$}
& \makecell{25/25\\$[0.863,1.000]$} \\

Iron
& \makecell{25/25\\$[0.863,1.000]$}
& \makecell{25/25\\$[0.863,1.000]$}
& \makecell{25/25\\$[0.863,1.000]$}
& \makecell{25/25\\$[0.863,1.000]$} \\

Diamond
& \makecell{11/25\\$[0.244,0.651]$}
& \makecell{17/25\\$[0.465,0.851]$}
& \makecell{18/25\\$[0.506,0.879]$}
& \makecell{23/25\\$[0.740,0.990]$} \\

Lava
& \makecell{19/25\\$[0.549,0.906]$}
& \makecell{25/25\\$[0.863,1.000]$}
& \makecell{25/25\\$[0.863,1.000]$}
& \makecell{25/25\\$[0.863,1.000]$} \\

Compass
& \makecell{17/25\\$[0.465,0.851]$}
& \makecell{25/25\\$[0.863,1.000]$}
& \makecell{9/25\\$[0.180,0.575]$}
& \makecell{25/25\\$[0.863,1.000]$} \\

\bottomrule
\end{tabular}

\caption{Task-success counts and exact two-sided 95\%
Clopper--Pearson confidence intervals over 25 trials per
method--task condition. S.L. denotes a skill library.}
\label{tab:goal-success-ci}
\end{table}

\section{Automaton State Counts}
\label{app:state-counts}

The product of a skill DFA and a specification DFA has the standard
worst-case upper bound $|Q_g||Q_c|$.
Table~\ref{tab:state-counts} reports the component sizes and the number
of reachable states in each composed automaton after unreachable
product states are removed.

\begin{table}[!ht]
\small
\centering
\setlength{\tabcolsep}{3.5pt}
\renewcommand{\arraystretch}{1.10}

\begin{tabular}{lccc}
\toprule
Metric
& \makecell{Diamond\\$\cap$ Sleep}
& \makecell{Mine\\$\cap$ Night}
& \makecell{Explore\\$\cap$ Biome} \\
\midrule

Skill DFA states
& 7
& N/A
& N/A \\

Specification DFA states
& 9
& 3
& 2 \\

Reachable product states
& 17
& 3
& 2 \\

\bottomrule
\end{tabular}

\caption{Component and reachable product-automaton sizes for the
evaluated skill--constraint compositions. The worst-case product size
is $|Q_g||Q_c|$; the reported product sizes exclude unreachable
states.}
\label{tab:state-counts}
\end{table}
\end{document}